\documentclass[twoside]{article}

\usepackage[accepted]{aistats2020}
\usepackage{microtype,marvosym}

\usepackage{tikz,pgfplots,pgfplotstable}
\usepackage{standalone}
\usepackage{extarrows}
\usepackage{subcaption}
\usepackage{amsmath,amsfonts,amssymb}
\usepackage[utf8]{inputenc}
\usepackage[T1]{fontenc}
\usepackage{lmodern}
\usepackage{nicefrac}
\usepackage{etoolbox}
\usepackage{enumitem}

\usepackage[super]{nth} 

\pgfplotsset{compat=newest}
\pgfplotsset{plot coordinates/math parser=false}
\pgfplotsset{select coords between index/.style 2 args={
    x filter/.code={
        \ifnum\coordindex<#1\fi
        \ifnum\coordindex>#2\fi
    }
}}
\pgfplotsset{ 
legend image code/.code={
\draw[mark repeat=2,mark phase=2]
plot coordinates {
(0cm,0cm)
(0.15cm,0cm)        
(0.3cm,0cm)         
};%
}}
\pgfset{
  foreach/parallel foreach/.style args={#1in#2via#3}{evaluate=#3 as #1 using {{#2}[#3-1]}},
}

\usetikzlibrary{arrows,shapes,plotmarks,pgfplots.colormaps,cd,positioning,angles,quotes}
\usetikzlibrary{external}
\usepgfplotslibrary{groupplots,fillbetween}
\tikzset{>=stealth'} 
\tikzstyle{graphnode} = 
   [circle,draw=black,minimum size=22pt,text centered,text
     width=22pt,inner sep=0pt] 
\tikzstyle{var}   =[graphnode,fill=white]
\tikzstyle{obs}   =[graphnode,fill=black,text=white]
\tikzstyle{fac}   =[rectangle,draw=black,fill=black!25,minimum size=5pt]
\tikzstyle{facprior} =[rectangle,draw=black,fill=black,text=white,minimum size=5pt]
\tikzstyle{edge}  =[draw=white,double=black,thick,-]
\tikzstyle{prior} =[rectangle, draw=black, fill=black, minimum size=
5pt, inner sep=0pt]
\tikzstyle{dirprior} = [circle, draw=black, fill=black, minimum
size=5pt, inner sep=0pt]

\AtBeginEnvironment{tikzcd}{\tikzexternaldisable}
\AtEndEnvironment{tikzcd}{\tikzexternalenable}

\pgfplotsset{
  every axis legend/.append style =
    {
      cells = { anchor = east },
      draw  = none
    },
}  
\pgfkeys{/pgfplots/mystyle/.style={
  xtick align = inside,
  ytick align = inside
  }}

\DeclareSymbolFont{stmry}{U}{stmry}{m}{n}
\DeclareMathSymbol\leftarrowtriangle\mathrel{stmry}{"5E}
\DeclareMathSymbol\rightarrowtriangle\mathrel{stmry}{"5F}
\DeclareMathSymbol\leftrightarrowtriangle\mathrel{stmry}{"5D}
\DeclareMathSymbol\obar\mathrel{stmry}{"3A}
\DeclareMathSymbol\otimes\mathrel{stmry}{"0F}
\DeclareMathSymbol\ominus\mathrel{stmry}{"17}
\DeclareMathSymbol\sslash\mathrel{stmry}{"0C}
\DeclareUnicodeCharacter{200E}{}

\usepackage{mathtools}
\usepackage{algorithm}
\usepackage{algpseudocode}
\algrenewcommand{\algorithmiccomment}[1]{\hfill {\footnotesize $\sslash$ #1}}
\algrenewcommand{\alglinenumber}[1]{\sf\scriptsize #1}

\makeatletter
\@addtoreset{algorithm}{chapter} 
\makeatother

\makeatletter
\def\therule{\makebox[\algorithmicindent][l]{\hspace*{.5em}\vrule height .75\baselineskip depth .25\baselineskip}}%

\newtoks\therules
\therules={}
\def\appendto#1#2{\expandafter#1\expandafter{\the#1#2}}
\def\gobblefirst#1{
  #1\expandafter\expandafter\expandafter{\expandafter\@gobble\the#1}}%
\def\LState{\State\unskip\the\therules}
\def\pushindent{\appendto\therules\therule}%
\def\popindent{\gobblefirst\therules}%
\def\printindent{\unskip\the\therules}%
\def\printandpush{\printindent\pushindent}%
\def\popandprint{\popindent\printindent}%

\algdef{SE}[WHILE]{While}{EndWhile}[1]
  {\printandpush\algorithmicwhile\ #1\ \algorithmicdo}
  {\popandprint\algorithmicend\ \algorithmicwhile}%
\algdef{SE}[FOR]{For}{EndFor}[1]
  {\printandpush\algorithmicfor\ #1\ \algorithmicdo}
  {\popandprint\algorithmicend\ \algorithmicfor}%
\algdef{S}[FOR]{ForAll}[1]
  {\printindent\algorithmicforall\ #1\ \algorithmicdo}%
\algdef{SE}[LOOP]{Loop}{EndLoop}
  {\printandpush\algorithmicloop}
  {\popandprint\algorithmicend\ \algorithmicloop}%
\algdef{SE}[REPEAT]{Repeat}{Until}
  {\printandpush\algorithmicrepeat}[1]
  {\popandprint\algorithmicuntil\ #1}%
\algdef{SE}[IF]{If}{EndIf}[1]
  {\printandpush\algorithmicif\ #1\ \algorithmicthen}
  {\popandprint\algorithmicend\ \algorithmicif}%
\algdef{C}[IF]{IF}{ElsIf}[1]
  {\popandprint\pushindent\algorithmicelse\ \algorithmicif\ #1\ \algorithmicthen}%
\algdef{Ce}[ELSE]{IF}{Else}{EndIf}
  {\popandprint\pushindent\algorithmicelse}%
\algdef{SE}[PROCEDURE]{Procedure}{EndProcedure}[2]
   {\printandpush\algorithmicprocedure\ \textproc{#1}\ifthenelse{\equal{#2}{}}{}{(#2)}}%
   {\popandprint\algorithmicend\ \algorithmicprocedure}%
\algdef{SE}[FUNCTION]{Function}{EndFunction}[2]
   {\printandpush\algorithmicfunction\ \textproc{#1}\ifthenelse{\equal{#2}{}}{}{(#2)}}%
   {\popandprint\algorithmicend\ \algorithmicfunction}%
\makeatother

\setlist{itemsep=3pt}

\usepackage{colonequals}
\newcommand{\ce}{\colonequals}

\usepackage{csquotes} 
\usepackage{hyperref}

\hypersetup{
    colorlinks,
    citecolor=black,
    filecolor=black,
    linkcolor=black,
    urlcolor=black
}
\usepackage[style=authoryear,
            maxcitenames=2,
            maxbibnames=5,
            firstinits=true,
            uniquelist=false,
            uniquename=false,
            backend=biber,
            natbib=true, 
            url=false,
            doi=false,
            eprint=false,
            isbn=false,
            dashed=false,
            hyperref=true]{biblatex}
\DeclareCiteCommand{\cite}
  {\usebibmacro{prenote}}
  {\usebibmacro{citeindex}%
   \printtext[bibhyperref]{\usebibmacro{cite}}}
  {\multicitedelim}
  {\usebibmacro{postnote}}

\DeclareCiteCommand*{\cite}
  {\usebibmacro{prenote}}
  {\usebibmacro{citeindex}%
   \printtext[bibhyperref]{\usebibmacro{citeyear}}}
  {\multicitedelim}
  {\usebibmacro{postnote}}

\DeclareCiteCommand{\parencite}[\mkbibparens]
  {\usebibmacro{prenote}}
  {\usebibmacro{citeindex}%
    \printtext[bibhyperref]{\usebibmacro{cite}}}
  {\multicitedelim}
  {\usebibmacro{postnote}}

\DeclareCiteCommand*{\parencite}[\mkbibparens]
  {\usebibmacro{prenote}}
  {\usebibmacro{citeindex}%
    \printtext[bibhyperref]{\usebibmacro{citeyear}}}
  {\multicitedelim}
  {\usebibmacro{postnote}}

\DeclareCiteCommand{\footcite}[\mkbibfootnote]
  {\usebibmacro{prenote}}
  {\usebibmacro{citeindex}%
  \printtext[bibhyperref]{ \usebibmacro{cite}}}
  {\multicitedelim}
  {\usebibmacro{postnote}}
  
\DeclareCiteCommand{\footcitetext}[\mkbibfootnotetext]
  {\usebibmacro{prenote}}
  {\usebibmacro{citeindex}%
   \printtext[bibhyperref]{\usebibmacro{cite}}}
  {\multicitedelim}
  {\usebibmacro{postnote}}

\DeclareCiteCommand{\textcite}
  {\boolfalse{cbx:parens}}
  {\usebibmacro{citeindex}%
   \printtext[bibhyperref]{\usebibmacro{textcite}}}
  {\ifbool{cbx:parens}
     {\bibcloseparen\global\boolfalse{cbx:parens}}
     {}%
   \multicitedelim}
  {\usebibmacro{textcite:postnote}}

\DeclareFieldFormat{titlecase}{\MakeSentenceCase*{#1}}

\newcommand{\gp}{\text{\textsc{gp}}}
\newcommand{\ep}{\text{\textsc{ep}}}
\newcommand{\ess}{\text{\sc ess}} 
\newcommand{\liness}{\text{\sc lin-ess}} 
\newcommand{\mcmc}{\text{\sc mcmc}} 
\newcommand{\smc}{\text{\sc smc}} 
\newcommand{\hdr}{\text{\sc hdr}} 

\newcommand{\pmin}{p_{\mathrm{min}}}
\newcommand{\hpmin}{\hat{p}_{\mathrm{min}}}

\newcommand{\ie}{i.e.,} %
\newcommand{\eg}{e.g.,} %
\newcommand{\g}{\,|\,} 
\renewcommand{\d}{\:d}  
\newcommand{\Exp}{\mathbb{E}}
\newcommand{\one}{\mathbf{1}}
\newcommand{\N}{\mathcal{N}} 
\renewcommand{\L}{\mathcal{L}}
\renewcommand{\Re}{\mathbb{R}}
\newcommand{\inv}{^{-1}} 
\newcommand{\Trans}{^{\intercal}} 
\newcommand{\dbyd}[2]{\frac{\operatorname{d}{#1}}{\operatorname{d}{#2}}}
\newcommand{\ddbyd}[3]{\frac{\operatorname{d}^2{#1}}{\operatorname{d}{#2}\operatorname{d}{#3}}}

\newcommand{\q}{\quad}
\renewcommand{\vec}{\boldsymbol} 
\renewcommand{\O}{\mathcal{O}} 

\newcommand{\bgamma}{\boldsymbol{\gamma}}
\newcommand{\bmu}{\boldsymbol{\mu}}
\newcommand{\bnu}{\boldsymbol{\nu}}

\renewcommand{\a}{\boldsymbol{\mathsf{a}}}
\renewcommand{\b}{\boldsymbol{\mathsf{b}}}
\newcommand{\f}{\boldsymbol{\mathsf{f}}}
\renewcommand{\u}{\boldsymbol{\mathsf{u}}}

\newcommand{\x}{\boldsymbol{\mathsf{x}}}
\newcommand{\z}{\boldsymbol{\mathsf{z}}}

\newcommand{\sA}{\boldsymbol{\mathsf{A}}}
\newcommand{\sL}{\boldsymbol{\mathsf{L}}}
\newcommand{\sM}{\boldsymbol{\mathsf{M}}}
\newcommand{\sX}{\boldsymbol{\mathsf{X}}}
\newcommand{\sSigma}{\boldsymbol{\mathsf{\Sigma}}}

\definecolor{dblu}{RGB}{0,0,130}
\definecolor{forestgreen}{RGB}{34, 139, 34}

\begin{document}

%

%

\twocolumn[

\aistatstitle{Integrals over Gaussians under Linear Domain Constraints}

\aistatsauthor{ Alexandra Gessner \And Oindrila Kanjilal \And Philipp Hennig}

\aistatsaddress{
\parbox[t]{\linewidth}
{\centering University of Tuebingen and\\ MPI for Intelligent Systems\\ T\"ubingen, Germany\\ \vskip0.4ex \href{mailto:agessner@tue.mpg.de}{\texttt{agessner@tue.mpg.de}}}
\And
\parbox[t]{\linewidth}
{\centering University of Tuebingen and\\ Technical University of Munich\\ Germany\\ \vskip0.4ex \href{mailto:oindrila.kanjilal@tum.de}{\texttt{oindrila.kanjilal@tum.de}}}
\And
\parbox[t]{\linewidth}
{\centering University of Tuebingen and\\ MPI for Intelligent Systems\\ T\"ubingen, Germany\\ \vskip0.4ex \href{mailto:ph@tue.mpg.de}{\texttt{ph@tue.mpg.de}}}
} 
]

\begin{abstract}
Integrals of linearly constrained multivariate Gaussian densities are a frequent problem in machine learning and statistics, arising in tasks like generalized linear models and Bayesian optimization.
Yet they are notoriously hard to compute, and to further complicate matters, the numerical values of such integrals may be very small.
We present an efficient black-box algorithm that exploits geometry for the estimation of integrals over a small, truncated Gaussian volume, and to simulate therefrom.
Our algorithm uses the Holmes-Diaconis-Ross (\hdr) method combined with an analytic version of elliptical slice sampling (\ess).
Adapted to the linear setting, \ess~allows for \emph{rejection-free} sampling, because intersections of ellipses and domain boundaries have closed-form solutions.
The key idea of \hdr~is to decompose the integral into easier-to-compute conditional probabilities by using a sequence of nested domains.
Remarkably, it allows for direct computation of the logarithm of the integral value and thus enables the computation of extremely small probability masses.
We demonstrate the effectiveness of our tailored combination of \hdr~and \ess~on high-dimensional integrals and on entropy search for Bayesian optimization.
\end{abstract}

\section{INTRODUCTION}

Multivariate Gaussian \emph{densities} are omnipresent in statistics and machine learning.
Yet, Gaussian \emph{probabilities} are hard to compute---they require solving an integral over a constrained Gaussian volume---owing to the intractability of the multivariate version of the Gaussian cumulative distribution function (\textsc{cdf}).
The probability mass that lies within a domain $\L\subset\Re^D$ restricted by $M$ linear constraints can be written as
\begin{equation}
Z = P(\x\in\L) = \int_{\Re^D} \prod_{m=1}^M \Theta \left(\a_m\Trans \x + b_m\right)\; \d\: \N (\x; 0, \one),
\label{eqn:integral}
\end{equation}
with the Heaviside step function $\Theta (x) = 1$ if $x>0$ and zero otherwise.
We take the integration measure to be a standard normal without loss of generality, because any correlated multivariate Gaussian can be whitened by linearly transforming the integration variable.

Gaussian models with linear domain constraints occur in a myriad of applications that span all disciplines of applied statistics and include biostatistics \citep{ThiebautJ2004}, medicine \citep{ChenC2007}, environmental sciences \citep{Wani2017}, robotics and control \citep{Fisac2018}, machine learning \citep{Su2016} and more. 
A common occurrence of this integral is in spatial statistics, such as Markov random fields \citep{BolinL2015}, the statistical modeling of spatial extreme events called max-stable processes \citep{HuserD2013, GentonYH2011}, or in modeling uncertainty regions for latent Gaussian models. An example for the latter is to find regions that are likely to exceed a given reference level, \eg~pollution levels in geostatistics and environmental monitoring \citep{BolinL2015}, or in climatology \citep{FrenchS2013}.
Another area where integrals like Eq.~\eqref{eqn:integral} are often encountered is in reliability analysis \citep{AuB2001b,Melchers2018,AndersenLR2018,StraubSBK2020}.
A key problem there is to estimate the probability of a rare event to occur (\eg~a flood) or for a mechanical system to enter a failure mode.\\
In machine learning, there are many Bayesian models in which linearly constrained multivariate normal distributions play a role, such as Gaussian processes under linear constraints \citep{LopezLoperaBDR2017, LopezLoperaJD2019, Agrell2019, DaVeigaM2012}, inference in graphical models \citep{MulgraveG2018}, multi-class Gaussian process classification \citep{RasmussenW2006}, ordinal and probit regression \citep{Lawrence2008, Ashford1970}, incomplete data classification \citep{Liao2007}, and Bayesian optimization \citep{HennigS2012, Wang2016}, to name a few.

This practical relevance has fed a slow-burn research effort in the integration of truncated Gaussians over decades \citep{Geweke1991, Genz1992, Joe1995, Vijverberg1997, Nomura2014}.
\citet{GassmannDS2002} and \citet{GenzB2009} provide comparisons and attest that the algorithm by \citet{Genz1992} provides the best accuracy across a wide range of test problems, which has made it a default choice in the literature.
Genz's method applies a sequence of transformations to transform the integration region to the unit cube $[0,1]^D$ and then solves the integral numerically using quasi-random integration points.
Other methods focus on specialized settings such as bivariate or trivariate Gaussian probabilities \citep{Genz2004, HayterL2013}, or on orthant probabilities \citep{MiwaHK2003, Craig2008, Nomura2016, HayterL2012}.
Yet, these methods are only feasible for at most a few tens of variables.
Only recent advances have targeted higher-dimensional integrals: \citet{AzzimontiG2017} study high-dimensional orthant probabilities and \citet{GentonKT2017} consider the special case where the structure of the covariance matrix allows for hierarchical decomposition to reduce computational complexity. \citet{Phinikettos2011} employ a combination of four variance reduction techniques to solve such integrals with Monte Carlo methods. \citet{Botev2016} constructs an exponential tilting of an importance sampling measure that builds on the method by \citet{Genz1992} and reports effectiveness for $D\lesssim 100$.
A different approach has been suggested by \citet{CunninghamHL2011}: They use expectation propagation to approximate the constrained normal integrand of Eq.~\eqref{eqn:integral} by a moment-matched multivariate normal density. This allows for fast integration, at the detriment of guarantees. Indeed, the authors report cases in which \ep~is far off the ground truth integral.

Closely related to integration is \emph{simulation} from linearly constrained Gaussians, yet these tasks have rarely been considered concurrently, except for \citet{Botev2016} who proposes an accept-reject sampler alongside the integration scheme. Earlier attempts employ Gibbs sampling \citep{Geweke1991}, or other Monte Carlo techniques \citep{CongCZ2017}.
\citet{KochB2019} recently introduced an algorithm for exact simulation from truncated Gaussians. Their method iteratively samples from transformed univariate truncated Gaussians that satisfy the box constraints.

In our work, we jointly address the sampling and the normalization problem for linearly constrained domains in a Gaussian space, making the following contributions:
\begin{itemize}[leftmargin=*, topsep=0pt, itemsep=1pt]
 \item We present an adapted version of elliptical slice sampling (\ess) which we call \liness~that allows for \emph{rejection-free} sampling from the linearly constrained domain $\L$. Its effectiveness is not compromised even if the probability mass of $\L$ is very small (cf.~Section \ref{sec:ess}).
  \item Based on the above \liness~algorithm, we introduce an efficient integrator for truncated Gaussians. It relies on a sequence of nested domains to decompose the integral into multiple, easier-to-solve, conditional probabilities. The method is an adapted version of the Holmes-Diaconis-Ross algorithm \citep{DiaconisH1995,Ross2012,KroeseTB2011} (cf.~Section \ref{sec:integration}). 
 \item With increasing dimension $D$, the integral value $Z$ can take extremely small values. \hdr~with a \liness~sampler allows to compute such integrals efficiently, and to even compute the logarithm of the integral. 
 \item With \liness, sampling is sufficiently efficient to also compute \emph{derivatives} of the probability with respect to the parameters of the Gaussian using expectations. 
\end{itemize}
We provide a \textsc{Python} implementation available at \url{https://github.com/alpiges/LinConGauss}.

\section{METHODS}
\label{sec:methods}

We first introduce an adapted version of elliptical slice sampling, \liness, which permits efficient sampling from a linearly constrained Gaussian domain of arbitrarily small mass once an initial sample within the domain is known. This routine is a special case of elliptical slice sampling that leverages the analytic tractability of intersections of ellipses and hyperplanes to speed up the \ess~loop.
\liness~acts at the back-end of the integration method, which is introduced in Section \ref{sec:integration}. 

For further consideration, it is convenient to write the linear constraints of Eq.~\eqref{eqn:integral} in vectorial form, $\sA\Trans \x + \b$, where $\sA\in\Re^{D\times M},\ \x\in\Re^D$, and $\b\in\Re^M$.
The integration domain $\L\subset\Re^D$ is given by the intersection of the region where all the $M$ constraints exceed zero.
For example, orthant probabilities of a correlated Gaussian $\N(\bmu, \sSigma)$ can be written in the form of Eq.~\eqref{eqn:integral} by using the transformation $\x =\sL\z + \bmu$, where $\sL$ is the Cholesky decomposition of $\sSigma$.
Typically, we expect $M\geq D$, \ie~there are at least as many linear constraints as dimensions. This is because if $M<D$, there exists a transformation of $\x$ such that $D-M$ dimensions can be integrated out in closed form, and an $M$-dimensional integral with $M$ constraints remains.
However, there are situations in which integrating out dimensions might be undesired. This is the case, e.g., when samples from the untransformed integrand are required.

\subsection{Sampling from truncated Gaussians}
\label{sec:ess}
Elliptical slice sampling (\ess) by \citet{MurrayAM2010} is a Markov chain Monte Carlo (\mcmc)~algorithm to draw samples from a posterior when the prior is a multivariate normal distribution $\N(\bmu, \sSigma)$.
Given an initial location $\x_0\in\Re^D$, an auxiliary vector $\bnu\sim\N(\bmu, \sSigma)$ is drawn to construct an ellipse $\x(\theta) = \x_0 \cos \theta + \bnu \sin\theta$ parameterized by the angle $\theta\in [0, 2\pi]$.
In the general case, the algorithm proceeds similarly to regular slice sampling \citep{Neal2003}, but on the angular domain.
A likelihood threshold is defined, and rejected proposals (in $\theta$) with likelihood values below the threshold are used to adapt the bracket $[\theta_\mathrm{min}, \theta_\mathrm{max}]$ to sample from, until a proposal is accepted that serves as new $\x_0$ (see \citet{MurrayAM2010} for details).

\ess~is designed for generic likelihood functions. The special form of the likelihood in Eq.~\eqref{eqn:integral} can be leveraged to significantly simplify the \ess~algorithm:
\begin{enumerate}[leftmargin=*, topsep=0pt, itemsep=1pt]
    \item The selector $\ell(\x) \ce \prod_{m=1}^M \Theta [\a_m\Trans\x + b_m]$  can take only the values 0 and 1. Hence there is no need for a likelihood threshold, the domain to sample from is always defined by $\ell(\x)=1$ for $\x(\theta)$ on the ellipse.
    \item The intersections between the ellipse and the linear constraints have closed-form solutions. The angular domain(s) to sample from can be constructed analytically, and \liness~is thus rejection-free. The typical bisection search of slice sampling becomes a simple analytic expression.
\end{enumerate}
\begin{figure}[h]
\begin{center}
  \includegraphics{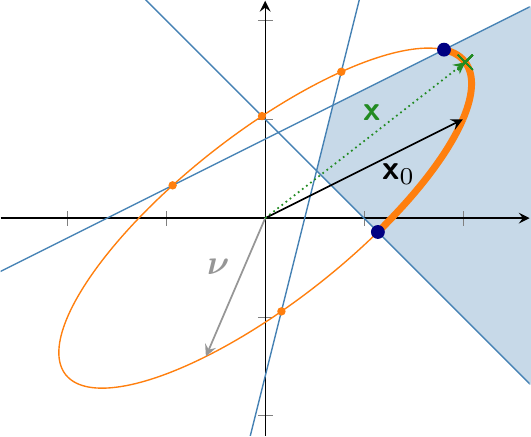}%
  \caption{Sampling from a constrained normal space using \ess. $\x_0$ is a previous sample from the domain $\L$ and, with the auxiliary $\bnu$, defines the ellipse. From all intersections of the ellipse and zero lines (or hyperplanes in higher dimensions), the \emph{active} intersections at the domain boundary are identified (\protect\tikz \protect\fill[dblu] (1ex,1ex) circle (0.4ex);). These define the slice from which a uniform sample is drawn ($\color{forestgreen}{\boldsymbol{\times}}$).}
  \label{fig:ellipse}
\end{center}
\end{figure}

With these simplifications of \ess, each sample from $\L$ requires exactly one auxiliary normal sample $\bnu\sim\N(0,\one)\in\Re^D$ and a scalar uniform sample $u\sim\text{Uniform}[0,1]$ to sample from the angular domain.
Fig. \ref{fig:ellipse} illustrates the process of drawing a sample from the domain of interest (blue shaded area) using our version of \ess. 
Given the two base vectors $\x_0\in\L$ and $\bnu$, the ellipse is parameterized by its angle $\theta\in [0,2\pi]$.
The intersections between the ellipse and the domain boundaries $\sA\Trans\x  + \b=\boldsymbol{0}$ can be expressed in closed form in terms of angles on the ellipse as solution to the set of equations $\sA\Trans(\x_0 \cos\theta + \bnu \sin\theta)  + \b= \boldsymbol{0}$.
For the $m^{\text{th}}$ constraint, this equation typically has either zero or two solutions,
\begin{equation}
\label{eqn:intersections}
    \theta_{m,1/2} = \pm\arccos \left(-\frac{b_m}{r}\right) + \arctan \left(\frac{\a_m\Trans \bnu}{r + \a_m\Trans \x_0}\right)
\end{equation}
with $r = \sqrt{(\a_m\Trans \x_0)^2 + (\a_m\Trans \bnu)^2}$.
A single solution occurs in the case of a tangential intersection, which is unlikely.
Not all intersection angles lie on the domain boundary and we need to identify those \emph{active} intersections where $\ell(\x(\theta))$ switches on or off.
To identify potentially multiple brackets, we sort the angles in increasing order and check for each of them if adding/subtracting a small $\Delta\theta$ causes a likelihood jump.
If there is no jump, the angle is discarded, otherwise the sign of the jump is stored (whether from 0 to 1 or the reverse), in order to know the direction of the relevant domain on the slice.
Pseudocode for \liness~can be found in Algorithm \ref{alg:dess} in the appendix.

The computational cost of drawing one sample on the ellipse is dominated by the $M$ inner products that need to be computed for the intersections, hence the complexity is $\O (MD)$.
This is comparable with standard \ess~for which drawing from a multivariate normal distribution is $\O(D^2)$, but the suppressed constant can be much smaller because there is no need to evaluate a likelihood function in \liness.
This version of \ess~is a rejection-free sampling method to sample from a truncated Gaussian of arbitrarily small mass---except that it requires an initial point within the domain from where to launch the Markov chain. How to obtain such a sample will be discussed in Section \ref{sec:subsetsim}.

\subsection{Computing Gaussian probabilities}
\label{sec:integration}

\subsubsection{The Holmes-Diaconis-Ross algorithm}
\label{sec:hdr}

The Holmes-Diaconis-Ross algorithm (\hdr) \citep{DiaconisH1995, Ross2012, KroeseTB2011} is a specialized method for constructing an unbiased estimator for probabilities of the form $P(\x\in\L)$ under an arbitrary prior measure $\x\sim p_0(\x)$ and a domain $\L=\{\x\ \text{ s.t. } f(\x)\geq 0\}$ with a deterministic function $f:\Re^D\mapsto\Re$. If this domain has very low probability mass, $P(\L)
$ is expensive to compute with direct Monte Carlo because most samples are rejected. \hdr~mitigates this by using a sequence of $T$ nested domains $\Re^D = \L_0 \supset \L_1 \supset \L_2 \supset ... \supset \L_T = \L$, s.t.~$\L_t = \bigcap_{i=1}^t \L_i$.
The probability mass of the domain of interest can be decomposed into a product of conditional probabilities,
\begin{equation}
\label{eqn:condprobs}
 Z = P(\L) = P(\L_0) \prod_{t=1}^{T} P(\L_{t} | \L_{t-1}).
\end{equation}
If each of the conditional probabilities $P(\L_{t+1} | \L_t)$ is closer to $\nicefrac{1}{2}$, they all require quadratically fewer samples, reducing the overall cost despite the linear increase in indidivual sampling problems. Noting that $P(\L_0) = 1$ and introducing the shorthand $\rho_t = P(\L_t | \L_{t-1})$, Eq.~\eqref{eqn:condprobs} can be written in logarithmic form as $\log Z= \sum_{t=1}^{T} \log\rho_t$.

\hdr~does not deal with the construction of these nested domains---a method to obtain them is discussed in Section~\ref{sec:subsetsim}.
For now, they are assumed to be given in terms of a decreasing sequence of positive scalar values $\{\gamma_1,\dots,\gamma_T\}$, where $\gamma_T = 0$.
Each shifted domain $\L_t$ can then be defined through its corresponding shift value $\gamma_t$.
In the general setting, this is $\L_t=\{\x\ \text{s.t.}\ f(\x)+\gamma_t\geq 0\}$; in our specific problem of linear constraints, $\x\in\L_t$ if $\ell_t (\x) = \prod_{m=1}^M \Theta(\a_m\Trans\x + b_m +\gamma_t) = 1$.
Any positive shift $\gamma_t$ thus induces a domain $\L_t$ that contains all domains $\L_{t'}$ with $\gamma_{t'}<\gamma_t$, and that engulfs a larger volume than $\L_{t'}$.
The $T^\mathrm{th}$ shift $\gamma_T = 0$ identifies $\L$ itself.

\begin{algorithm}[t]
\caption{The Holmes-Diaconis-Ross algorithm applied to linearly constrained Gaussians}
\begin{algorithmic}[1]
\Procedure{HDR}{$\sA, \b, \{\gamma_1,\dots,\gamma_T\}, N$}
\LState $\sX \sim \N (0, \one)$ \Comment{$N$ samples}
\LState $\log Z = 0$ \Comment{initialize log integral value}
\For{$t=1\dots T$}
    \LState $\L_t = \{\x: \min_m (\a_m\Trans \x_n + b_m) + \gamma_t > 0\}_{n=1}^N$
    \LState    \Comment{find samples inside current nesting}
    \LState $\log Z \leftarrow \log Z + \log (\# (\sX\in \L_t)) - \log N $
    \LState choose $\x_0 \in \L_t$
    \LState $\sX \leftarrow$ \Call{LinESS}{$\sA, \b+\gamma_t, N, \x_0$} 
    \LState    \Comment{draw new samples from constrained domain}
\EndFor
\LState\Return $\log Z$
\EndProcedure
\end{algorithmic}
\label{alg:hdr}
\end{algorithm}

Given the shift sequence $\{\gamma_1,\dots,\gamma_T\}$, the \hdr~algorithm proceeds as follows:
Initially, $N$ samples are drawn from $\L_0$, the integration measure, in our case a standard normal.
$\L_0$ corresponds to $\gamma_0 = \infty$ which is ignored in the sequence.
The conditional probability $\rho_1 = P(\L_1\g \L_0)$ is estimated as the fraction of samples from $\L_0$ that also fall into $\L_1$.
To estimate the subsequent conditional probabilities $\rho_t$ for $t>1$ as the fraction of samples from $\L_{t-1}$ falling into $\L_t$, standard \hdr~uses an \mcmc~sampler to simulate from $\L_{t-1}$.
If the sequence of nestings is chosen well and initial seeds in the domain $\L_{t-1}$ are known, these samplers achieve a high acceptance rate.
This procedure is repeated until $t=T$.
With the estimated conditional probabilities $\hat{\rho}_t$, the estimator for the probability mass is then
\begin{equation}
    \log \hat{Z} = \sum_{t=1}^T \log \hat{\rho}_t.
\end{equation}
In our adapted version of \hdr, the \liness~algorithm (cf. Section \ref{sec:ess}) comes into play, which achieves a 100\% acceptance rate for simulating from the nested domains.
In order to simulate rejection-free from $\L_t$, \liness~requires an initial sample from the domain $\L_t$, which is obtained from the previous iteration of the algorithm.
Every location sampled requires evaluating the linear constraints, hence the cost for each subset in \hdr~is $\O(NMD)$.
Pseudocode for this algorithm is shown in Algorithm~\ref{alg:hdr}, where \textsc{LinESS} is a call to the \liness~sampler (cf.~Section~\ref{sec:ess} and Algorithm~\ref{alg:dess} in the appendix) that simulates from the linearly constrained domain.

\subsubsection{Obtaining nested domains}
\label{sec:subsetsim}

As the final missing ingredient, the \hdr~algorithm requires a sequence of nested domains or level sets defined by positive shifts $\gamma_t$, $t=1,\dots, T$.
In theory, the nested domains should ideally have conditional probabilities of $\rho_t = \nicefrac{1}{2}$ $\forall t$ (then each nesting improves the precision by one bit).
Yet, in a more practical consideration, the computational overhead for constructing the nested domains should also be small.
In practice, the shift sequence is often chosen in an ad hoc way, hoping that conditional probabilities are large enough to enable a decently accurate estimation via \hdr~\citep{KanjilalM2015}.
This is not straightforward and requires problem-specific knowledge.

\begin{figure*}[t]
\begin{center}
  \includegraphics{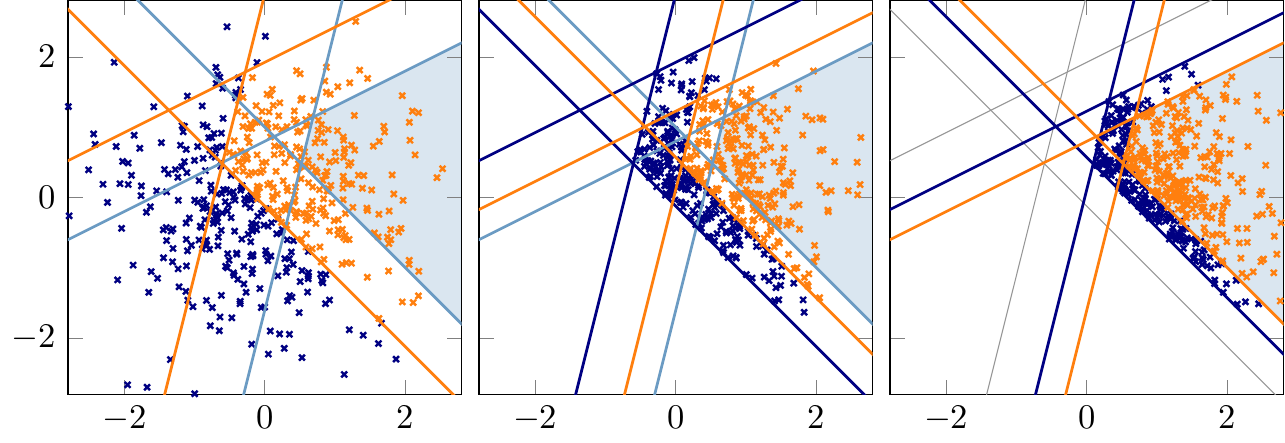}%
  \caption{Finding the level sets in subset simulation for linear constraints. \emph{Left:} Draw standard normal samples and find the shift $\gamma_1$ for which a fraction $\rho$ of the samples lie inside the new domain (orange lines); \emph{center:} Use \liness~to draw samples from the subsequent domain defined by $\gamma_1$ (now in dark blue) and find $\gamma_2$ (orange lines) similarly; \emph{right:} Proceed until the domain of interest (shaded area) is reached. Details in text.}
  \label{fig:subsets}
\end{center}
\end{figure*}

We suggest to construct the nestings via subset simulation \citep{AuB2001b} which is very similar to \hdr.
It only differs in that the conditional probabilities $\rho_t$ are fixed a priori to a value $\rho$, and then the shift values $\gamma_t$ are computed such that a fraction $\rho$ of the $N$ samples drawn from $\L_{t-1}$ falls into the subsequent domain $\L_t$.\\
The construction of the nested domains is depicted in Fig. \ref{fig:subsets}.
To find the shifts, $N$ samples are drawn from the integration measure initially (cf.~Fig. \ref{fig:subsets}, left).
Then the first (and largest) shift $\gamma_1$ is determined such that a fraction $\rho$ of the samples fall into the domain $\L_1$.
This is achieved by computing for each sample by how much the linear constraints would need to be shifted to encompass the sample.
For the subsequent shifts, $N$ samples are simulated from the current domain $\L_t$, and the next shift $\gamma_t$ is again set s.t. $\lfloor N\rho\rfloor$ samples fall into the next domain $\L_{t+1}$ (Fig. \ref{fig:subsets}, center).
This requires an initial sample from $\L_t$ to launch the \liness~sampler, which is obtained from the samples gathered in the previous nesting $\L_{t-1}$ that also lie in $\L_t$, while all other samples are discarded to reduce dependencies.
This nesting procedure is repeated until more than $\lfloor N\rho\rfloor$ samples fall into the domain of interest $\L$ (cf. Fig. \ref{fig:subsets}, right).
We set $\rho=\nicefrac{1}{2}$ to maximize the entropy of the binary distribution over whether samples fall in- or outside the next nested domain, yet in reliability analysis a common choice is $\rho=0.1$ \citep{AuB2001a}, which has the advantage of requiring less nestings (to the detriment of more samples).
Pseudocode can be found in Algorithm~\ref{alg:subset} in the appendix.

In fact, subset simulation itself also permits the estimation of the integral $Z$, without appealing to \hdr: Since the subsets are constructed such that the conditional probabilities take a predefined value, the estimator for the integral is $\hat{Z}_{\text{ss}} = \rho^{T-1} \rho_T$ where $\rho_T = P(\L_{T}|\L_{T-1}) \in [\rho, 1]$ is the conditional probability for the last domain.
For $\rho=\nicefrac{1}{2}$ the number of nestings is roughly the negative binary logarithm of the integral estimator $T\approx -\log_2 \hat{Z}_{\text{ss}}$ (cf. Fig.~\ref{fig:shifts}). 
The main reason not to rely on subset simulation alone is that its estimator $\hat{Z}_{\text{ss}}$ is biased, because the samples are both used to construct the domains and to estimate $Z$.
We thus use \hdr~for the integral estimation and subset simulation for the construction of the level sets.

Both subset simulation and \hdr~are instances of a wider class of so-called \emph{multilevel splitting} methods which are related to \emph{sequential Monte Carlo} (\smc) in that they are concerned with simulating from a sequence of probability distributions. \smc~methods (aka. \emph{particle filters}) were conceived for online inference in state space models, but can be extended to non-Markovian latent variable models \citep{NaessethLS2019}. In this form, \smc~methods have gained popularity for the estimation of rare events \citep{DelMoralDJ2006, BectLV2017, CerouDFG2013}.

\subsubsection{Derivatives of Gaussian probabilities}
\label{sec:derivatives}
Many applications (e.g. Bayesian optimization, see below) additionally require \emph{derivatives} of the Gaussian probability w.r.t. to parameters $\lambda$ of the integration measure or the linear constraints. The absence of such derivatives in classic quadrature sub-routines (such as from \citet{Genz1992}) has thus sometimes been mentioned as an argument against them \citep[e.g.][]{CunninghamHL2011}). Our method allows to efficiently compute such derivatives, because it can produce samples. This leverages the classic result that derivatives of exponential families with respect to their parameters can be computed from expectations of the sufficient statistics.
To do so, it is advantageous to rephrase Eq.~\eqref{eqn:integral} as the integral over a \emph{correlated} Gaussian with mean $\bmu$ and covariance matrix $\sSigma$ with axis-aligned constraints (or constraints that are independent of $\lambda$).
The derivatives w.r.t.~a parameter $\lambda$ can then be expressed as an expected value,
\begin{equation}
\label{eqn:derivatives}
    \dbyd{Z}{\lambda} = \Exp \left[ \dbyd{\log \N(\x;\bmu, \sSigma)}{\lambda}\right],
\end{equation}
where the expectation is taken with respect to the transformed integrand Eq.~\eqref{eqn:integral}.
Since \liness~permits us to simulate from the integrand of Eq.~\eqref{eqn:integral}, derivatives can be estimated via expectations.
We demonstrate in Section~\ref{sec:boes} that this is a lot more efficient than finite differences, which requires $Z$ to be estimated twice, and at considerably higher accuracy.

\section{EXPERIMENTS}
\label{experiments}

To shed light on the interplay of subset simulation, \hdr, and \liness, we consider a 500-dimensional synthetic integration problem with a closed-form solution.
Further 1000-d integrals can be found in Section~\ref{sec:synthexp_appendix}.
We then turn to Bayesian optimization and demonstrate our algorithm's ability to estimate derivatives.

\subsection{Synthetic experiments}
\label{sec:synthetic}
As an initial integration problem we consider axis-aligned constraints in a 500-dimensional space.
Since this task amounts to computing the mass of a shifted orthant under a standard normal distribution, it allows comparison to an exact analytic answer.
The goal of this setup is two-fold: 1) to demonstrate that our method can compute small Gaussian probabilities to high accuracy, and 2) to explore configurations for the construction of nested domains using subset simulation.
The domain is defined by $\ell(\x) = \prod_{d=1}^D \Theta(x_d + 1)$.
The true mass of this domain is $3.07\cdot 10^{-38}=2^{-124.6}$.
Estimating this integral na\"ively by sampling from the Gaussian would require of the order of $10^{38}$ samples for one to fall into the domain of interest. With a standard library like \texttt{numpy.random.randn}, this would take about $10^{15}$ ages of the universe.

\begin{figure}[t!]\centering
  \includegraphics{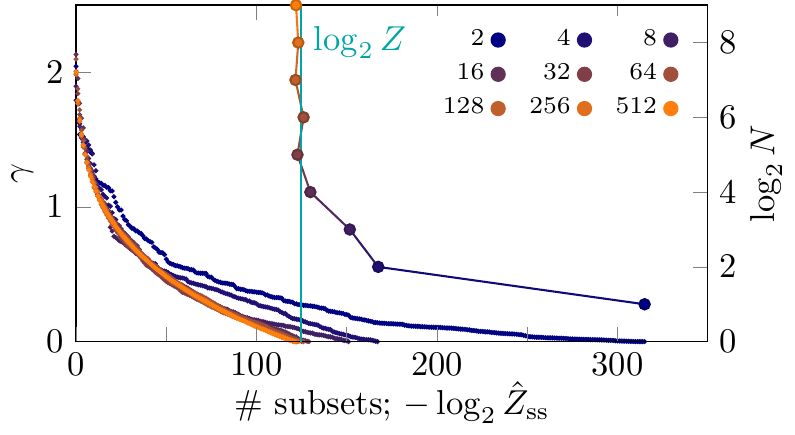}%
  \caption{Shift values $\gamma$ against number of subsets $T$ for different sample size per nesting $N$ (small dots). The connected dots show $-\log_2 \hat{Z}_{\text{ss}}$ vs. $\log_2 N$. The ground truth is indicated by the vertical line. This plot emphasizes the connection between $T$ and $-\log_2 Z$ for $\rho=\nicefrac{1}{2}$ (see text for details).}
  \label{fig:shifts}
\end{figure}

\paragraph{Subset simulation}
First, we compute the shift sequence $\{\gamma_1, \dots,\gamma_T\}$ using subset simulation for various numbers of samples $N$ per subset and a fixed conditional probability of $\rho=\nicefrac{1}{2}$.
Since the contributing factor of each nesting is $\rho=\nicefrac{1}{2}$, the integral estimate is roughly $2^{-T}$ for our choice of $\rho$ (cf. Section \ref{sec:subsetsim}).
The relation between the number of subsets $T$ and the estimated integral value $\hat{Z}_{\text{ss}}$ is visualized in Fig.~\ref{fig:shifts}. It shows the sequences of shift values for increasing sample sizes and the resulting integral estimate $\log_2 \hat{Z}_{\text{ss}}$.
The $T^{\mathrm{th}}$ nesting has shift value $\gamma=0$ and is the only subset with a conditional probability that deviates from the chosen value of $\rho$, yet $T$ is a good indicator for the value of the negative binary logarithm of the estimated integral. Hence we use the same axis to display the number of subsets and $-\log_2 \hat{Z}_{\text{ss}}$.
The plot highlights the bias of subset simulation: For small sample sizes, e.g.~$N=2,4,8$, the integral is severely underestimated.
This bias is caused by the dependency of the subset construction method on the samples themselves:
Since we are using a \mcmc~method for simulating from the current domain, samples are correlated and do not fall into the \emph{true} next subset with probability exactly $\rho$.
This is why we only accept every $10^\mathrm{th}$ sample to diminish this effect when constructing the subsets.
For the subsequent \hdr~simulation, we accepted every second sample from the \ess~procedure.\\
We choose powers of 2 for the number of samples per subset and observe that as of 16 samples per subset, the subset sequence is good enough to be handed to \hdr~for more accurate and unbiased estimation.
This low requirement of 16 samples per nesting also means that subset simulation is a low-cost preparation for \hdr, and causes only minor computational overhead.

\begin{figure*}[h]\centering
  \includegraphics{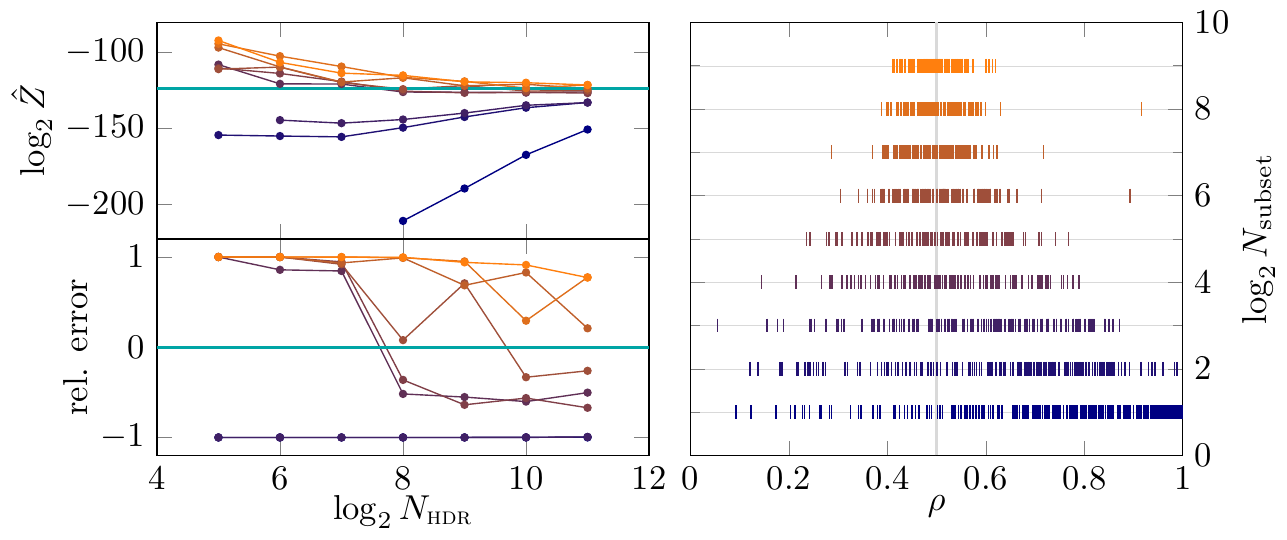}%
    \caption{\emph{Left:} \hdr~integral estimates for different subset sequences (same color coding as in Fig.~\ref{fig:shifts}) for $2^5$ to $2^{11}$ samples, \emph{top:} compared to the binary logarithm of the ground truth (horizontal line), and \emph{bottom:} the relative error. \emph{Right:} Conditional probabilities obtained by \hdr~for the same subset sequences, where $\rho=\nicefrac{1}{2}$ was chosen for the construction of the subsets (vertical line).}
  \label{fig:hdr}
\end{figure*}

\paragraph{Holmes-Diaconis-Ross}
Fig.~\ref{fig:hdr} shows the results achieved by \hdr~for the nine subset sequences obtained with $2^1$ to $2^9$ samples per subset and for different numbers of samples per nesting for \hdr.
The top left panel of Fig.~\ref{fig:hdr} shows the binary logarithm of the \hdr~integral estimator.
The bad performance for the subsets created with 2, 4, or 8 samples per nesting indicates that a good nesting sequence is essential for the effectiveness of \hdr, but also that such a sequence can be found using only about 16 samples per subset (this is thus the number used for all subsequent experiments).
The bottom left panel displays the relative error of the \hdr~estimator.
It is to bear in mind that the relative error is $\nicefrac{9}{11}$ if the estimator is one order of magnitude off, indicating that \hdr~achieves the right order of magnitude with a relatively low sample demand.
The right panel of Fig.~\ref{fig:hdr} shows the values for the conditional probabilities found by \hdr, using $2^{11}$ samples per subdomain.
If subset simulation were perfectly reliable, these should ideally be $\rho=\nicefrac{1}{2}$.
The plot confirms that, with $N\geq 16$, all conditional probabilities found by \hdr~are far from 0 and 1, warranting the efficiency of \hdr.

\subsection{Bayesian optimization}
\label{sec:boes}
Bayesian optimization is a sample-efficient approach to global optimization of expensive-to-evaluate black-box functions (see \citet{ShahriariSWAF2016} for a review).
A surrogate over the objective function $f(\x)$ serves to build a utility function and ultimately derive a policy to determine the next query point.
Information-based utilities are directly concerned with the posterior distribution over the minimizer, $\pmin (\x\g\mathcal{D})$, where $\mathcal{D} = \{\x_n, f(\x_n)\}_{n=1}^N$ summarizes previous evaluations of $f$.
Entropy search \citep{HennigS2012} seeks to evaluate the objective function at the location that bears the most information about the minimizer.
The expression $\pmin (\x\g\mathcal{D})$ is an infinite-dimensional integral itself, but for practical purposes, it can be discretized considering the distribution over so-called \emph{representer points}.
The probability of the $i^{\text{th}}$ representer point to be the minimum can be approximated as
\begin{equation}
    \hpmin (\x_i) = \int\! \d\f\; \N (\f, \bmu, \sSigma) \prod_{j\neq i} \Theta (f(\x_j) - f(\x_i)),
\label{eqn:pmin}
\end{equation}
where $\bmu$ and $\sSigma$ are the posterior mean and covariance of the Gaussian process over $f$, respectively.
Clearly, this is a linearly constrained Gaussian integral in the form of Eq.~\eqref{eqn:integral} which has to be solved for all $N_R$ representer points. Eq. \eqref{eqn:pmin} is stated in matrix form in the appendix Section~\ref{sec:appendix_boes}.
The original paper and implementation uses expectation propagation (\ep) to approximate this integral.

\begin{figure}[h]\centering
  \includegraphics{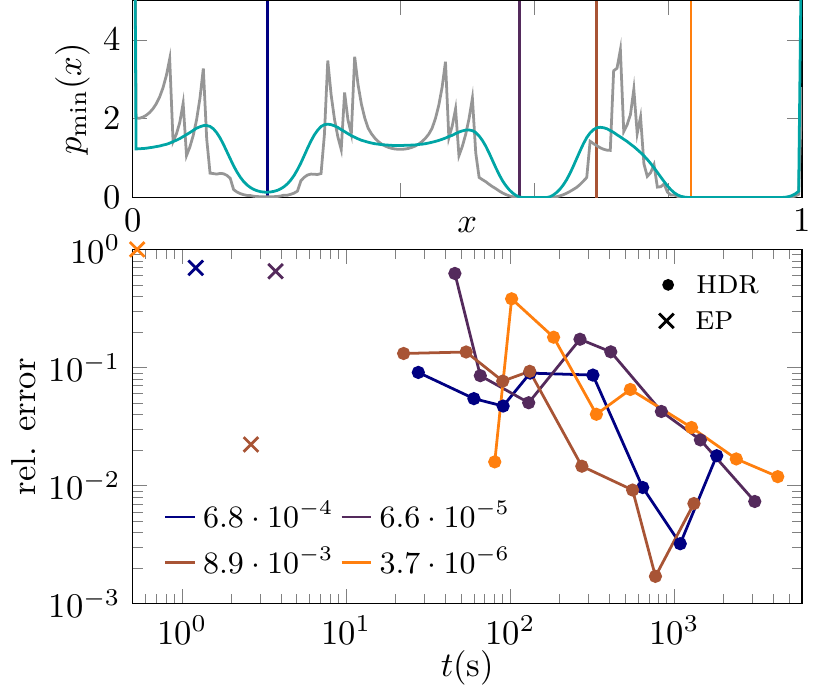}%
  \caption{\emph{Top:} Probability for $x$ to be the minimum, estimated via Thompson sampling (blue), and \ep~(gray). Vertical lines indicate locations at which we run \hdr. \emph{Bottom:} Absolute relative error by \ep~and \hdr~against \textsc{cpu} time at the locations indicated above. Each \hdr~sequence shown uses $2^6$ to $2^{13}$ samples per nesting. The smaller $\hpmin$, the longer takes the \hdr~run, since there are more subsets to traverse.}
  \label{fig:bo-es}
\end{figure}

\paragraph{Probability of minimum}
For our experiment, we consider the one-dimensional Forrester function \citep{Forrester2007} with three initial evaluations.
The top plot in Fig.~\ref{fig:bo-es} shows the ground truth distribution over the minimum obtained by Thompson sampling, \ie~drawing samples from the discretized posterior \gp~and recording their respective minimum, and the approximation over this distribution obtained by \ep.
It is apparent that \ep~fails to accurately represent $\hpmin$.
For \hdr, we consider four locations (indicated by the vertical lines) and show that while it takes longer to compute, the estimate obtained by \hdr~converges to the true solution (see bottom plot of Fig.~\ref{fig:bo-es}).
In the experiment we use 200 representer points---which is an unusually high number for a 1-d problem---to show that our method can deal with integrals of that dimension.
Also note that we are reporting \textsc{cpu} time, which means that due to automatic parallelization in \textsc{Python} the wall clock time is considerably lower.

\paragraph{Derivatives}
Entropy search requires derivatives of Eq.~\eqref{eqn:pmin} to construct a first-order approximation of the predictive information gain from evaluating at a new location $\x_\star$.
We can estimate derivatives using expectations (cf. Section~\ref{sec:derivatives} and \ref{sec:appendix_boes}).
Initially we choose 5 representer points to validate the approach of computing derivatives via moments against finite differences.
The latter requires estimating $\hpmin$ at very high accuracy and has thus a high sample demand even in this low-dimensional setting, for which we employ both rejection sampling and \hdr.
The derivatives computed via moments from rejection sampling and \liness~take $0.7\%$ of the time required to get a similar accuracy with finite differences.
Unsurprisingly, rejection sampling is faster in this case, with $\hpmin(\x_i)\approx\nicefrac{1}{4}$, i.e.~only $\sim\nicefrac{3}{4}$ of the samples from the posterior over $f$ need to be discarded to obtain independent draws that have their minimum at $\x_i$.
\liness~only outperforms rejection sampling at higher rejection rates common to higher-dimensional problems.\\
Therefore, we also consider 20 representer points, which corresponds to a 20-d linearly constrained space to sample from.
In this setting, we consider a location of low probability, with $\hpmin=1.6\cdot 10^{-4}$, which renders an estimation via finite differences impossible and highly disfavors rejection sampling even for computing the moments.
\liness, however, enables us to estimate the gradient of the normal distribution w.r.t. its mean and covariance matrix with a relative standard deviation on the 2-norm of the order of $10^{-2}$ using $5\cdot 10^5$ samples and an average \textsc{cpu} time of 325\,s for a problem that was previously unfeasible.
A badly conditioned covariance matrix in Eq.~\eqref{eqn:derivatives} deteriorates runtime (which is already apparent in the considered case) since it requires estimating moments at very high accuracy to compensate for numerical errors.

\subsection{Constrained samples}
\label{sec:constrained_samples}

\begin{figure}[t]\centering
  \includegraphics{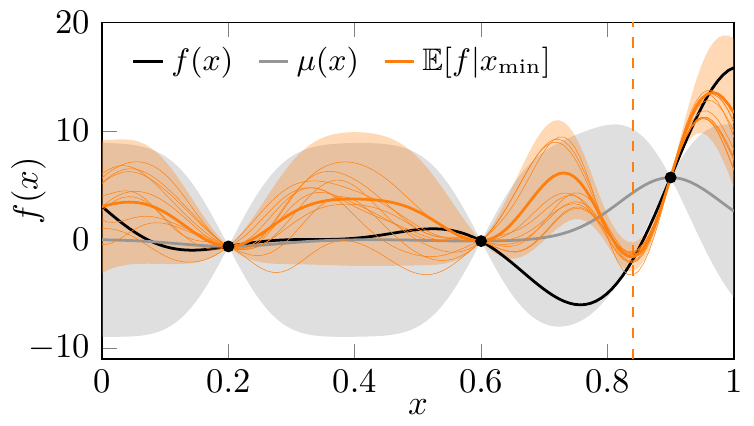}%
  \caption{The Forrester function (black), the posterior \gp~given three evaluations (gray), and the posterior distribution over $f$ conditioned on the minimum being located at where the vertical line indicates (orange), each with the $2\sigma$ confidence interval shaded. The latter has been obtained from drawing $10^5$ samples using \liness, 10 of which are shown (thin orange lines).}
  \label{fig:xmin}
\end{figure}

We emphasize that \liness~allows to draw samples from linearly constrained Gaussians \emph{without rejection}.
In the Gaussian process setting, this permits to efficiently draw samples that are subject to linear restrictions \citep{Agrell2019, LopezLoperaBDR2017, DaVeigaM2012}.
In particular, the time required for sampling is essentially independent of the probability mass of the domain of interest.
This probability mass only affects the precomputation required to find an initial sample in the domain for \liness~(cf. Section~\ref{sec:subsetsim}).
Since this can be achieved with $\sim\! 16$ samples per subset (cf. Section~\ref{sec:synthetic}), this initial runtime is typically negligible compared to the actual sampling.
Fig.~\ref{fig:xmin} displays the posterior distribution of a \gp~conditioned on the location of the minimum from the Bayesian optimization context, estimated from \liness~samples.
This distribution is required in predictive entropy search \citep{HernandezHG2014}---a reformulation of the original entropy search---where it is approximated by imposing several related constraints (\eg~on the derivatives at the minimizer $\x_{\min}$).
The probability for the given location to be the minimizer is $\lesssim 10^{-6}$, which renders direct sampling virtually impossible.
The unaltered \ess~algorithm fails on this problem due to the domain selector---a binary likelihood.

\section{CONCLUSIONS}

We have introduced a black-box algorithm that computes Gaussian probabilities (i.e.~the integral over linearly constrained Gaussian densities) with high numerical precision, even if the integration domain is of high dimensionality and the probability to be computed is very small. This was achieved by adapting two separate pieces of existing prior art and carefully matching them to the problem domain: We designed a special version of elliptical slice sampling that takes explicit advantage of the linearly-constrained Gaussian setting, and used it as an internal step of the \hdr~algorithm. We showed that, because this algorithm can not just compute integrals but also produces samples from the nestings alongside, it also permits the evaluation of derivatives of the integral with respect to the parameters of the measure. One current limitation is that, because our algorithm was designed to be unbiased, it has comparably high computational cost (but also superior numerical precision) over alternatives like expectation propagation. This problem could be mitigated if one is willing to accept unbiasedness and thus reuse samples. Furthermore, both \hdr~and \liness~are highly parallelizable (as opposed to \ep) and thus offer margin for implementational improvement.

\subsection*{Acknowledgements}
AG and PH gratefully acknowledge financial support by the European Research Council through ERC StG Action 757275 / PANAMA; the DFG Cluster of Excellence ``Machine Learning - New Perspectives for Science'', EXC 2064/1, project number 390727645; the German Federal Ministry of Education and Research (BMBF) through the Tübingen AI Center (FKZ: 01IS18039A); and funds from the Ministry of Science, Research and Arts of the State of Baden-Württemberg. 
The work was carried out while OK was at the University of Tuebingen, funded by the German Research Foundation (Research Unit 1735).
OK also acknowledges financial support through the Alexander von Humboldt Foundation. 
AG is grateful to the International Max Planck Research School for Intelligent Systems (IMPRS-IS) for support.
 
\printbibliography

\appendix
\pagebreak

\twocolumn[
\begin{center}
\thispagestyle{empty}
\hsize\textwidth
  \linewidth\hsize \toptitlebar {\centering
  {\Large\bfseries Supplementary Material \par}
  \vspace{5pt}
  \large{\textbf{Integrals over Gaussians under Linear Domain Constraints}}}
 \bottomtitlebar \vskip 0.2in plus 1fil minus 0.1in
\end{center}
]

\section{ALGORITHMS}

\begin{minipage}{\textwidth}
\begin{algorithm}[H]
\caption{Elliptical slice sampling for a linearly constrained standard normal distribution}
\begin{algorithmic}[1]
\Procedure{LinESS}{$\sA, \b, N, \x_0$} 
\LState \textbf{ensure} \textbf{all}($\a_m\Trans \x_0 + b_m > 0\ \forall m$) \Comment{initial vector needs to be in domain}
\LState $\sX = [\ ]$ \Comment{initialize sample array}
\For{n = 1,\dots, N}
\LState $\bnu \sim \N (0,\one)$
\LState $\x(\theta) = \x_0 \cos \theta + \bnu \sin \theta$ \Comment{construct ellipse}
\LState $\boldsymbol{\theta} \leftarrow \text{\textbf{sort}}(\{\theta_{j, 1/2}\}_{j=1}^M)$ s.t. $\a_j\Trans (\x_0 \cos \theta_{j,1/2} + \bnu \sin \theta_{j,1/2}) = 0$ 
\Comment{$2M$ intersections, Eq. \eqref{eqn:intersections}}\vspace{-0.35em}
\LState $\boldsymbol{\theta}_\mathrm{act} \leftarrow \{[\theta^\mathrm{min}_l, \theta^\mathrm{max}_l]\}_{l=1}^L\q \text{s.t.}\ \ell (x(\theta^\mathrm{min/max}_l + d\theta)) - \ell (x(\theta^\mathrm{min/max}_l - d\theta)) = \pm 1$ \Comment{Set brackets} \vspace{-0.2em}
\LState $u\sim[0,1]\cdot \sum_l^L (\theta^\mathrm{max}_l - \theta^\mathrm{min}_l)$\vspace{-0.1em}
\LState $\theta_u \leftarrow$ transform $u$ to angle in bracket
\LState $\sX[n] \leftarrow \x(\theta_u)$ \Comment{update sample array}
\LState $\x_0 \leftarrow \x(\theta_u)$ \Comment{set new initial vector}
\EndFor
\LState\Return $\sX$
\EndProcedure
\end{algorithmic}
\label{alg:dess}
\end{algorithm}

\begin{algorithm}[H]
\caption{Subset simulation for linear constraints}
\begin{algorithmic}[1]
\Procedure{SubsetSim}{$\sA, \b, N, \rho=\frac{1}{2}$}
\LState $\sX \sim \N (0, \one)$ \Comment{$N$ initial samples}
\LState $\gamma, \hat{\rho} =$ \Call{FindShift}{$\rho$, $\sX$, $\sA, \b$}             \Comment{find new shift value}
\LState $\log Z = \log \hat{\rho}$ \Comment{record the integral}
\While{$\gamma>0$}
    \LState $\sX \leftarrow$ \Call{LinESS}{$\sA, \b+\gamma, N, \x_0$} 
        \Comment{draw new samples from new constrained domain}
    \LState $\gamma, \hat{\rho} \leftarrow$ \Call{FindShift}{$\rho$, $\sX$, $\sA, \b$} 
        \Comment{find new shift value}
    \LState $\log Z \leftarrow \log Z +  \log \hat{\rho}$
        \Comment{Update integral with new conditional probability}
\EndWhile
\LState\Return $\log Z$, shift sequence
\EndProcedure
\vspace{2ex}
\Function{FindShift}{$\rho$, $\sX$, $\sA, \b$} 
    \Comment{find shift s.t. a fraction $\rho$ of $\sX$ fall into the resulting domain.}
\LState $\bgamma \leftarrow$ \textsc{sort}$(-{\min_m (\a_m\Trans \x_n + b_m)}_{n=1}^N)$
    \Comment{sort shifts in ascending order}
\LState $\gamma \leftarrow (\bgamma[\lfloor\rho N\rfloor] + \bgamma[\lfloor\rho N\rfloor+1])/2$ \Comment{Find shift s.t. $\rho N$ samples lie in the domain}
\LState $\hat{\rho} \leftarrow (\# \sX\ \text{inside})/N$ \Comment{true fraction could deviate from $\rho$}
\LState\Return $\gamma$, $\hat{\rho}$
\EndFunction
\end{algorithmic}
\label{alg:subset}
\end{algorithm}
\end{minipage}

\newpage
\phantom{text needed?}
\newpage
\section{DETAILS ON EXPERIMENTS}

\subsection{Synthetic experiments}
\label{sec:synthexp_appendix}
\paragraph{1000-d integrals}
We further consider three similar synthetic integrals over orthants of 1000-d correlated Gaussians with a fixed mean and a randomly drawn covariance matrix.
Table \ref{tbl:exp1000d} shows the mean and std.~dev.~of the binary logarithm of the integral estimator averaged over five runs of \hdr~using $2^8$ samples per nesting for integration, as well as the average \textsc{cpu} time\footnote{On 6 \textsc{cpu}s, the wall clock time was $\sim$20\,min per run.}.

\begin{table}[h]\centering
\caption{Integrals of Gaussian orthants in 1000-d}
\vspace{-1mm}
\begin{tabular}{cccc}
  \#  & $\langle\log_2\hat{Z}\rangle$ & std.~dev. &  $t_{\text{\sc cpu}}$[$10^3$s] \\
  \hline
  1 & $-162.35$ & $4.27$ & $8.86$\\
  2 & $-160.54$ & $2.09$ & $7.40$\\
  3 & $-157.62$ & $3.19$ & $7.64$\\
\end{tabular}
\label{tbl:exp1000d}
\end{table}

\subsection{Bayesian optimization}
\label{sec:appendix_boes}

\paragraph{Probability of minimum}
After having chosen $N_R$ representer points, the approximate probability for $\x_i, i=1,\dots,N_R$ to be the minimum, Eq.~\eqref{eqn:pmin} can be rephrased in terms of Eq.~\eqref{eqn:integral} by writing the $N_R-1$ linear constraints in matrix form. This $(N_R -1)\times N_R$ matrix is a $(N_R -1)\times (N_R-1)$ identity matrix with a vector of $-\one$ added in the i$^{th}$ column,

\begin{equation*}
    \sM = 
    \begin{bmatrix} 
        \one_{(i-1)\times(i-1)} & {\color{dblu} -\one_{i-1}} & \vec{0}_{(i-1)\times(N_R-i)}\\
        \vec{0}_{(N_R-i)\times(i-1)} & {\color{dblu} -\one_{N_R-i}} & \one_{(N_R-i)\times(N_R-i)}\\
    \end{bmatrix}.
\end{equation*}
Then the objective Eq.~\eqref{eqn:pmin} can be written as
\begin{equation*}
\begin{aligned}
    \hpmin (\x_i) &= \int \N (\f, \bmu, \sSigma) \prod_{j\neq i}^{N_R} \Theta ([\sM \f]_j) \d\f\\
    &= \int \N (\u, \vec{0}, \one) \prod_{j\neq i}^{N_R} \Theta \left(\left[\sM \left(\sSigma^{\nicefrac{1}{2}}\u + \bmu\right)\right]_j \right) \d\u\\
\end{aligned}
\end{equation*}
where we have done the substitution $\u=\sSigma^{-\nicefrac{1}{2}} (\f - \bmu)$, and hence $\f = \sSigma^{\nicefrac{1}{2}}\u + \bmu$.
Writing the constraints in matrix form as in Section~\ref{sec:methods}, $\sA\Trans = \sM \sSigma^{\nicefrac{1}{2}}$ and $\b = \sM\bmu$.

\paragraph{Derivatives}
In order to compute a first-order approximation to the objective function in entropy search, we need the derivatives of $\hpmin$ w.r.t. the parameters $\bmu$ and $\sSigma$.
The algorithm requires the following derivative, where $\lambda = \{\bmu, \sSigma\}$,

\begin{equation*}
\begin{aligned}
    &\dbyd{}{\lambda} \log \pmin 
    \approx \frac{1}{\hpmin} \dbyd{\hpmin}{\lambda}\\
    &= \frac{1}{\hpmin} \int \d\f\ \dbyd{\N (\f, \bmu, \sSigma)}{\lambda} \prod_{j\neq i}^{N_R} \Theta ([\sM \f]_j )\\
    &= \frac{1}{\hpmin} \Exp \left[\dbyd{\log \N (\f,\bmu,\sSigma)}{\lambda}\right],
\end{aligned}
\end{equation*}

using $\dbyd{\N (\f, \bmu, \sSigma)}{\lambda} = \N (\f, \bmu, \sSigma) \dbyd{\log \N (\f, \bmu, \sSigma)}{\lambda}$.
Hence all we need is to compute the derivatives of the log normal distribution w.r.t. its parameters, and the expected values thereof w.r.t. the integrand.
The required derivatives are
\begin{equation*}
    \dbyd{\log \N (\f, \bmu, \sSigma)}{\mu_i} 
    = \left[ \sSigma\inv (\f-\bmu) \right]_i,
\end{equation*}
\begin{equation*}
    \begin{aligned}
    \dbyd{\log \N (\f, \bmu, \sSigma)}{\sSigma_{ij}}
    = \frac{1}{2}  \left[ \sSigma\inv (\f-\bmu)(\f-\bmu)\Trans \sSigma\inv - \sSigma\inv \right]_{ij}
    \end{aligned}
\end{equation*}
and the second derivative
\begin{equation*}
    \begin{aligned}
    &\ddbyd{\N (\f, \bmu, \sSigma)}{\mu_i}{\mu_j}\\
    &= \N (\f, \bmu, \sSigma) \left(\left[\sSigma\inv (\f-\bmu) (\f-\bmu)\Trans \sSigma\inv - \sSigma\inv\right]_{ij} \right)
    \end{aligned}
\end{equation*}

Hence we only need $\Exp_{\pmin} [(\f-\bmu)]$ and $\Exp_{\pmin} [(\f-\bmu)(\f-\bmu)\Trans]$ to compute the following gradients,

\begin{equation*}
    \dbyd{\log \pmin}{\mu_i} \approx \frac{1}{\hpmin} \Exp_{\hpmin} \left[ \left[ \sSigma\inv (\f-\bmu) \right]_i \right],
\end{equation*}
\begin{equation*}
    \begin{aligned}
    &\dbyd{\log \pmin}{\sSigma_{ij}} \approx\\
    &\frac{1}{\hpmin} \Exp_{\hpmin} \left[\frac{1}{2}  \left[ \sSigma\inv (\f-\bmu)(\f-\bmu)\Trans \sSigma\inv - \sSigma\inv \right]_{ij} \right],
    \end{aligned}
\end{equation*}
and the Hessian w.r.t. $\bmu$,
\begin{equation*}
    \ddbyd{\log \pmin}{\mu_i}{\mu_j}
    = 2 \dbyd{\log \hpmin}{\sSigma_{ij}} - \dbyd{\log \pmin}{\mu_i} \dbyd{\log \pmin}{\mu_j}.
\end{equation*}

\end{document}